\documentclass[journal,twoside,web]{ieeecolor}
\usepackage{amsmath}
\usepackage{amssymb}
\usepackage{bm}
\usepackage{mathtools}
\usepackage{verbatim}
\usepackage{tablefootnote}

\usepackage{algorithm}
\usepackage{algorithmic}

\def\etal{\emph{et al}.}

\usepackage{bbold}
\usepackage{bbm}
\usepackage{booktabs} %
\usepackage{array}  %
\usepackage[tight,normalsize,sf,SF]{subfigure}
\usepackage{multirow}
\usepackage{tabularx}
\usepackage{url}
\usepackage{color}
\usepackage{float}
\usepackage{gensymb}
\usepackage{cite}

\usepackage{tabu}
\usepackage{dcolumn}

\usepackage{graphicx}

\usepackage[pagebackref=true,breaklinks=true,letterpaper=true,colorlinks,bookmarks=false]{hyperref}

\usepackage{generic}
\usepackage{cite}
\usepackage{amsmath,amssymb,amsfonts}
\usepackage{algorithmic}
\usepackage{graphicx}
\usepackage{textcomp}
\def\BibTeX{{\rm B\kern-.05em{\sc i\kern-.025em b}\kern-.08em
    T\kern-.1667em\lower.7ex\hbox{E}\kern-.125emX}}
\begin{document}
\title{A Discrepancy Aware Framework for Robust Anomaly Detection}
\author{Yuxuan Cai,  Dingkang Liang, \IEEEmembership{Graduate Student Member, IEEE}, Dongliang Luo, Xinwei He, Xin Yang, \IEEEmembership{Member, IEEE}, and Xiang Bai, \IEEEmembership{Senior Member, IEEE}
\thanks{Manuscript submitted 19 February 2023; revised 23 August 2023; accepted 13 September 2023. This work was supported in part by the Young Scientists Fund of the National Natural Science Foundation of China under Grant 62302188, and in part by the National Science Fund for Distinguished Young Scholars of China under Grant 62225603. Paper no. TII-23-0558. (Corresponding author: Xinwei He)}
\thanks{Yuxuan Cai, Dingkang Liang, and Xiang Bai are with the School of Artificial Intelligence and Automation, Huazhong University of Science and Technology, Wuhan 430074, China (e-mail: cyx\_hust@hust.edu.cn; dkliang@hust.edu.cn; xbai@hust.edu.cn)}
\thanks{Dongliang Luo and Xin Yang are with the School of Electronic Information and Communications, Huazhong University of Science and Technology, Wuhan 430074, China (e-mail: ldl@hust.edu.cn; xinyang2014@hust.edu.cn)}
\thanks{Xinwei He is with the College of Informatics, Huazhong Agriculture University, Wuhan 430070, China (e-mail: xwhe@mail.hzau.edu.cn)}}

\maketitle

\begin{abstract}
Defect detection is a critical research area in artificial intelligence. Recently, synthetic data-based self-supervised learning has shown \textcolor{black}{great potential on this task.} Although many sophisticated synthesizing strategies exist, little research has been done to investigate the robustness of models when faced with different strategies. In this paper, we focus on this issue and find that existing methods are highly sensitive to them.
To alleviate this issue, we present a Discrepancy Aware Framework (DAF), which demonstrates robust performance consistently with simple and cheap strategies across different anomaly detection benchmarks. 
We hypothesize that the high sensitivity to synthetic data of existing self-supervised methods arises from their heavy reliance on the visual appearance of synthetic data during decoding. In contrast, our method leverages an appearance-agnostic cue to guide the decoder in identifying defects, thereby alleviating its reliance on synthetic appearance. To this end, \textcolor{black}{inspired by existing knowledge distillation methods,} we employ a teacher-student network, which is trained based on synthesized outliers, to compute the discrepancy map as the cue.
Extensive experiments on two challenging datasets prove the robustness of our method. Under the simple synthesis strategies, it outperforms existing methods by a large margin. Furthermore, it also achieves the state-of-the-art localization performance. Code is available at: \href{https://github.com/caiyuxuan1120/DAF}{https://github.com/caiyuxuan1120/DAF}.
\end{abstract}

\IEEEpeerreviewmaketitle

\section{Introduction}\label{sec:introduction}

\IEEEPARstart{I}{mage} anomaly detection plays an important role in many safety-critical areas, e.g., industrial manufacturing \textcolor{black}{systems}~\cite{Ni2021AttentionNF,sem}, surveillance systems~\cite{Gong2019MemorizingNT}, and medical image analysis~\cite{caiyus}.
\textcolor{black}{However, in these areas, acquiring sufficient high-quality anomaly images is generally difficult or even impossible. This limitation hinders the effectiveness of training deep-learning models through supervised methods.}
As a result, it has induced \textcolor{black}{growing research interest}~\cite{bergmann2020uninformed,STPM,deng2022anomaly} in exploring approaches \textcolor{black}{that focus on training anomaly detection models solely on normal images.}
\textcolor{black}{Nevertheless, the lack of anomaly data during training poses significant challenges in extracting discriminative features for unseen anomaly data during inference.}

\begin{figure}[!t]
    \centering
    \includegraphics[width=1.0\columnwidth]{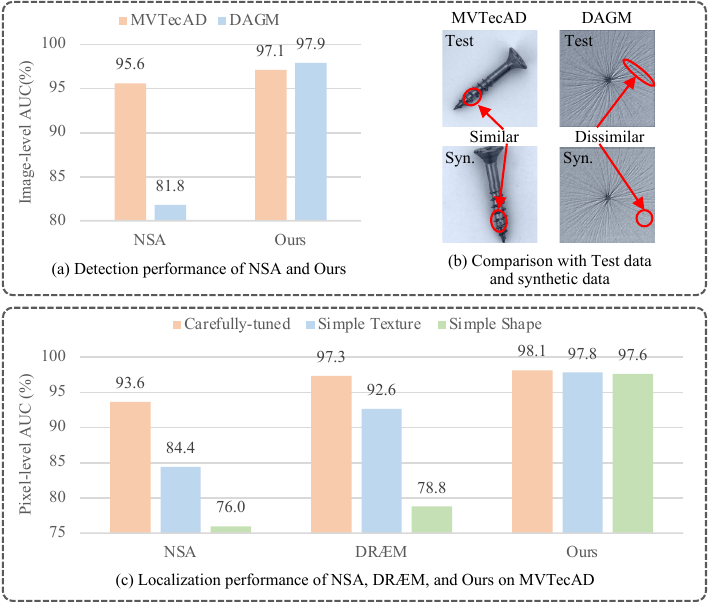}
    \caption{\textcolor{black}{(a) Detection performance comparison. (b) Real \emph{vs.} Synthetic images. (c) Localization performance under diverse synthetic strategies.}}
\label{figure1:overview}
\end{figure}

\begin{figure}[!t]
    \centering
    \includegraphics[width=1.0\columnwidth]{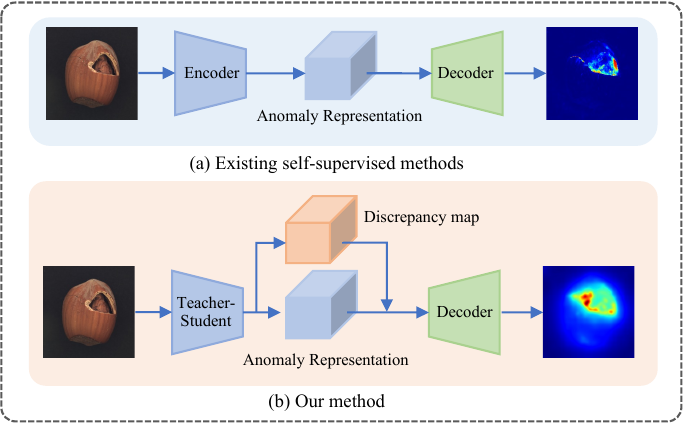}
    \caption{\textcolor{black}{Overview of existing self-supervised methods and ours.}}
\label{figure2:Ex-our-pipe}
\end{figure}

Recently, some researchers have assumed that models trained on normal data may fail to reconstruct anomaly patterns, \textcolor{black}{and have proposed identifying anomaly regions based on reconstruction failures.} However, these reconstruction-based methods may be easily misled in practice because of the identical shortcut issue~\cite{You2022AUM}. Another promising research direction is based on self-supervision~\cite{cutpaste,Zavrtanik_2021_ICCV,Schlter2022NaturalSA}. Generally, the typical framework follows an encoder-decoder architecture (Fig.~\ref{figure2:Ex-our-pipe}(a)). \textcolor{black}{Normal data is first distorted} to generate realistic and diverse outlier data. \textcolor{black}{Subsequently, the framework is trained} to differentiate normal and synthesized abnormal images. Such a strategy of exposing models to the synthesized outliers has demonstrated better empirical results, dominating current research in anomaly detection. 
\textcolor{black}{The success of these models heavily relies on generating diverse and close-to-real anomaly images, and much effort has been endeavored to improve anomaly synthesis strategies.}
For instance, DRÆM~\cite{Zavrtanik_2021_ICCV} uses Perlin noise to generate irregular shapes, simulating the shape of real anomalies. 
NSA~\cite{Schlter2022NaturalSA} integrates Poisson image editing~\cite{perez2003poisson} to eliminate \textcolor{black}{discontinuous borders of anomalous patterns}, making the anomalies more natural. 

Despite great success, few attempts have been made to study the robustness of current methods to different synthesis strategies. However, it has been noted~\cite{Schlter2022NaturalSA} in the community that the decoder tends to \textcolor{black}{overfit to the} synthetic anomaly appearance during the training phase. \textcolor{black}{As a result,  the decision boundary tends to generalize poorly to real anomalies during inference}. Moreover, the anomaly patterns generally show large variances across datasets.
Therefore, the synthesis strategy customized for one dataset may not \textcolor{black}{be well-suited to} another. As shown in Fig.~\ref{figure1:overview}(b), a carefully-tuned synthesis strategy for MVTecAD~\cite{bergmann2019mvtec} generates unnatural synthetic data for DAGM~\cite{wieler2007weakly}, which may help explain the dramatic performance degradation \textcolor{black}{of NSA~\cite{Schlter2022NaturalSA} on DAGM} (Fig.~\ref{figure1:overview}(a)).
Besides, our preliminary experiments on two representative self-supervision-based methods, \emph{i.e.}, NSA~\cite{Schlter2022NaturalSA} and DRÆM~\cite{Zavrtanik_2021_ICCV}, also indicate their high sensitivity to different synthesis strategies (Fig.~\ref{figure1:overview}(c)). 

\textcolor{black}{To tackle the above issues,} in this paper, we present a Discrepancy Aware framework (DAF) (Fig.~\ref{figure2:Ex-our-pipe}(b)), which can maintain strong performance consistently across various existing anomaly synthesis techniques. The core idea behind our method is releasing the decoder from the constraint of \textcolor{black}{the} synthetic anomaly appearance. To accomplish this, we leverage the appearance-agnostic discrepancy map derived from a teacher-student network as guidance for the decoder.
\textcolor{black}{Since the discrepancy map is computed based on extracted high-level features, it is less affected by the synthetic appearance. Moreover, by processing the discrepancy directly, the decoder will focus on discriminating the normal and non-normal regions.}
\textcolor{black}{During inference, given an anomaly image, the teacher and the student will demonstrate discrepancy in every non-normal region, while the segmentation probability map generated by the decoder will display a clear decision boundary for identifying anomalies.}

To the best of our knowledge, we are the first to investigate the robustness of current frameworks to different anomaly synthesis techniques. Compared with existing self-supervised methods, our method has the following desired properties. First, our framework incorporates the teacher-student network into the \textcolor{black}{self-supervised paradigm, enhancing} its capacity to produce discrepant features for anomaly regions. 
Second, our method encodes non-normal regions derived from the discrepancy map rather than \textcolor{black}{the non-normal appearance. This approach} reduces \textcolor{black}{overfitting} problems for the decoder during training, thus eliminating the heavy reliance on carefully-tuned anomaly synthesis techniques.
As shown in Fig.~\ref{figure1:overview}(c), our method reaches strong performance even with a simple synthesis strategy. Besides, our method also achieves significant performance improvements over existing methods, surpassing them by a large margin in terms of localization capability on MVTecAD~\cite{bergmann2019mvtec}. It also achieves \textcolor{black}{the state-of-the-art} detection performance on DAGM~\cite{wieler2007weakly}. 

In summary, the main contributions of this paper are as follows: 1) We introduce a simple and robust self-supervised framework named \textbf{DAF} for image anomaly detection and localization, which eliminates the practical need for the \textcolor{black}{complicated tuning steps for synthesis.} 2) We propose to combine the teacher-student network with the \textcolor{black}{self-supervised paradigm}, which utilizes abundant synthesized anomaly images to learn the discrepancy features, alleviating the \textcolor{black}{overfitting problem to the non-normal appearance.}

\section{Related Work} \label{sec:relatedwork}
\textcolor{black}{Early approaches~\cite{Hocenski2006ImprovedCE,Li2011DefectII} to image anomaly detection} typically work by first extracting the feature descriptors or statistical information and then calculating the anomaly scores. 

Recently, with remarkable progress in deep learning, image anomaly detection based on deep learning has become a dominant direction. Below we mainly review the deep learning-based approaches, which can be roughly grouped into three categories: reconstruction-based, self-supervision-based, and knowledge distillation-based approaches.

\subsection{\textcolor{black}{Reconstruction-based approaches}}
These approaches~\cite{liu2020towards,venkataramanan2020attention} assume that anomalies are difficult to be reconstructed by models trained only on normal images. Thus, the anomaly regions can be spotted by examining regions with larger reconstruction errors. In these methods, autoencoders are frequently adopted as reconstruction models. For instance, Baur \etal~\cite{baur2018deep} introduce deep spatial autoencoding architectures, which are trained by a pixel-wise reconstruction loss and an adversarial loss to improve the construction quality. 
Besides, generative adversarial networks~\cite{Yang2021AnAF, schlegl2019f} are also attractive models for reconstructing the input image. However, the assumption behind the reconstruction methods \textcolor{black}{might not always be valid}, as neural networks sometimes generalize to anomalies well and yield good reconstruction results.

\subsection{\textcolor{black}{Self-supervision-based approaches}}
Driven by the success of self-supervised learning in visual representation learning~\cite{He2021MaskedAA,He2019MomentumCF}, approaches under this paradigm have emerged rapidly and advanced the state-of-the-art. 
For instance, 
~\cite{zavrtanik2021reconstruction} trains an autoencoder to reconstruct \textcolor{black}{the masked image to the original one first,} \textcolor{black}{and then uses the reconstruction error} as the anomaly score. Different from ~\cite{zavrtanik2021reconstruction}, which constructs masks on images, SSPCAB~\cite{sspcab} applies the idea of masking \textcolor{black}{in convolution blocks.} In this way, it can be integrated into any CNN architecture.

Recent works~\cite{Zavrtanik_2021_ICCV,Schlter2022NaturalSA} prove that anomaly detection benefits from synthesized defects that are close to real ones. These works \textcolor{black}{carefully design} synthetic data strategies, then \textcolor{black}{train} segmentation models such as U-Net~\cite{ronneberger2015u} for pixel-level prediction, ensuring \textcolor{black}{that} the segmentation model can learn a suitable decision boundary \textcolor{black}{between} normal and abnormal regions. However, the segmentation model is likely to be ineffective \textcolor{black}{when there is a significant difference between the distribution of synthetic defects and actual ones.} Even though DRÆM~\cite{Zavrtanik_2021_ICCV} tries to prevent the model from \textcolor{black}{overfitting} to the synthetic data by introducing anomaly-free reconstruction, \textcolor{black}{as mentioned above,} the reconstruction model will also fail to restore \textcolor{black}{the anomalous region to normal one} during inference if the anomaly manifold is unseen during training. 
\textcolor{black}{Moreover, SPD~\cite{SPD} shifts its focus to self-supervised pre-training. Specifically, it proposes a novel augmentation strategy to encourage models to be locally sensitive, making the representations more suitable for the defect detection task.}

\subsection{\textcolor{black}{Knowledge distillation-based approaches} }
This group of approaches detects anomalies by reflecting images to different \textcolor{black}{representation spaces}, assuming that the representations of normal regions in different spaces are identical while those of abnormal regions will differ. 
The teacher-student framework is adopted to achieve the different-space reflection. 
For instance, prior works~\cite{bergmann2020uninformed,multiresolution,STPM} train the student network to mimic the pre-trained teacher on normal images. As a result, the student and the teacher tend to hold consistency on normal regions and display discrepancies on anomalies.
In the STAD~\cite{bergmann2020uninformed} framework, multiple students with the same structure are trained to regress the teacher network. The discrepancies between the teacher and students, along with the variance of students, are adopted to represent the anomaly score. MKDAD~\cite{multiresolution} and STPM~\cite{STPM} propose distilling \textcolor{black}{features} from various layers of the teacher to the corresponding layers of the student. RDAD~\cite{deng2022anomaly} utilizes reverse distillation to prohibit comparable anomaly representations \textcolor{black}{in different feature spaces.}
\textcolor{black}{SSMRKD~\cite{SSMRKD} further incorporates reverse distillation with the masking strategy and constructs a two-stage framework. In the first stage, it uses reverse distillation to construct a reconstruction network, enabling the student model to accurately reconstruct normal patterns. In the second stage,} the masking strategy is applied to enhance the sensitivity of the reconstruction model \textcolor{black}{to} anomalies, enabling \textcolor{black}{effective identification and classification of abnormal patterns.}

However, all the aforementioned knowledge distillation-based methods \textcolor{black}{assume} training models \textcolor{black}{using} clean data. Recently, SoftPatch~\cite{Softpatch} has addressed the scenario where the training data is contaminated with real defect data. They propose a patch-level denoising strategy to improve \textcolor{black}{the robustness of the model}.

Different from previous knowledge distillation-based methods, our method incorporates the teacher-student model into a synthetic data-based self-supervised framework. It is designed to distinguish between normal patterns and abnormal ones. This objective enables our method to establish a discriminative decision boundary. Furthermore, unlike previous arts that focus on using the discrepancy for defect localization, our method leverages the teacher-student model \textcolor{black}{to incorporate an appearance-agnostic cue into the self-supervised framework,} thereby improving its robustness to synthetic anomaly images.

\begin{table}[]\color{black}
    \centering
    \caption{\textcolor{black}{Symbol Description}}
    \begin{tabular}{cc|l}
        \toprule
         No. & Symbol &  Description \\
        \midrule
         1 & $T$ & The Teacher Network \\
         2 & $S$ & The Student Network  \\
         3 & $Seg$ & The Segmentation Head\\
         4 & $Aux_i$ & The $i\text{-}th$ Auxiliary Head \\
         5 & $f_{t}^i$ & Representations extracted by the teacher\\
         6 & $f_{s}^i$ & Representations extracted by the student\\
         7 & $\overline{M}$ & The Discrepancy Map \\
         8 & $M_S$ & The Segmentation Probability Map \\
         9 & $M_{Score}$ & The Anomaly Score Map \\
         10 & $CutP$ & The synthesis strategy in CutPaste~\cite{cutpaste} \\
         11 & $DRA$ & The synthesis strategy in DRÆM~\cite{Zavrtanik_2021_ICCV} \\
         12 & $NSA_B$ & The synthesis strategy in NSA~\cite{Schlter2022NaturalSA} \\
         \bottomrule
    \end{tabular}
    \label{tab:symboldescrip}
\end{table}

\section{Our Method}\label{sec:method} 

\subsection{Overview}

\begin{figure*}[!t]
    \centering
    \includegraphics[width=2.0\columnwidth]{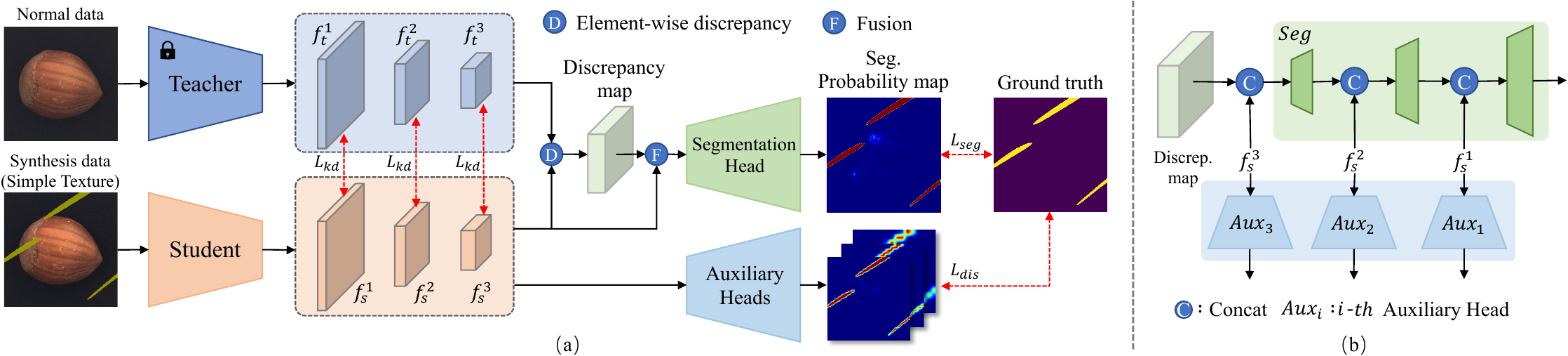}
    \caption{\textcolor{black}{(a) Overview of our method. (b) The fusion process.}}
\label{fig:pipeline}
\end{figure*}

\textcolor{black}{The framework consists of} a teacher-student network, a segmentation decoder, and a series of auxiliary heads, as shown in Fig.~\ref{fig:pipeline}.
\textcolor{black}{For clarity and ease of reference, Table ~\ref{tab:symboldescrip} summarizes essential symbols and their corresponding descriptions utilized in the following text.}
Given an input image, the teacher-student network ($T\text{-}S$) is expected to demonstrate discrepancies on non-normal regions. To accomplish this, the student is trained to maintain consistency with the teacher on normal regions during training.
The discrepancy maps yielded by the $T\text{-}S$ are subsequently fed into the segmentation head, along with representations of synthesized data, to localize anomalies.
Finally, auxiliary heads are employed to further supervise the student model to derive more discriminative representations for anomalous regions.

\subsection{Teacher-Student Network} 

In practice, anomalies occur in various, sometimes unconstrained formats. 
Some can be easily discerned by their distinct textures or colors from normal patterns, while others may require contextual information to be localized.
\textcolor{black}{Therefore, both low-level and high-level representations are vital for accurately localizing anomalies in such challenging scenarios.} In our design, the teacher-student framework follows a multi-scale knowledge distillation paradigm to represent anomalies \textcolor{black}{at different levels of granularity.}

\textcolor{black}{Following existing knowledge distillation methods~\cite{bergmann2020uninformed, STPM}, we adopt a powerful convolutional neural network, pre-trained on a large-scale dataset ImageNet~\cite{Deng2009ImageNetAL}, to initialize the teacher. As for the student, we choose the same architecture but initialize it randomly. Note that any off-the-shelf pre-trained networks could be adopted. Here we use ResNet18~\cite{he2016deep}, following STPM~\cite{STPM}. For the \textcolor{black}{multi-scale knowledge distillation}, we \textcolor{black}{transfer} the knowledge of three-stage features from the teacher to the student.} The distillation process is elaborated as follows.

\subsubsection{Training strategy} Given an anomaly-free training set $\mathcal{D} = \{I_m\}_{m=1}^{N}$, we first distort each $I_m \in \mathcal{D}$ to obtain \textcolor{black}{the} synthetic anomaly image $P_m$ using predefined anomaly strategies. 
The teacher network takes the anomaly-free image $I_m$ as input, while the student network takes the corresponding synthetic anomaly image $P_m$ as input. During training, we keep the teacher frozen and train the student to mimic the responses of the teacher on normal regions.
 
Following previous works~\cite{STPM, multiresolution}, cosine similarity is applied to measure the consistency between the representations of \textcolor{black}{the teacher and the student.} The cosine similarity loss is defined as:
\begin{equation}
  \textcolor{black}{L_{cos} = \sum_{i=1}^{3} (1 - \frac{1}{N_{neg}} \sum_{\Omega_{neg}}  \frac{V_{t}^{i} \cdot V_{s}^{i}}{||V_{t}^{i}|| \cdot ||V_{s}^{i}||})\textcolor{black}{\,,}}
  \label{eq:cossimilarity-loss}
\end{equation}
where \textcolor{black}{$i$ denotes the $i\text{-}th$ stage.} $\Omega_{neg}$ represents the normal \textcolor{black}{regions} and \textcolor{black}{$N_{neg}$ indicates the total number of pixels in normal regions.} \textcolor{black}{$V_{t}^{i}$ and $V_{s}^{i}$ represent the feature vectors yielded by the teacher and the student, respectively.}

However, \textcolor{black}{the cosine similarity only constrains each feature vector in isolation}, disregarding local contextual information. %
Thus we further introduce structural similarity ($SSIM$) loss for compensation by considering neighboring vectors.
The structural similarity is formulated as:
\begin{equation}
  \textcolor{black}{SSIM(p, q) = \frac{(2\mu_p\mu_q+\lambda_1)(2\sigma_{pq}+\lambda_2)}{(\mu_p^2+\mu_q^2+\lambda_1)(\sigma_p^2+\sigma_q^2+\lambda_2)}\textcolor{black}{\,,}}
  \label{eq:ssim}
\end{equation}
where $\mu$ and $\sigma$ represent the mean and variance, and $\lambda_1$ and $\lambda_2$ are adopted for numerical stability\footnote{$\lambda_1$, $\lambda_2$ are empirically set to 1e-1, 9e-4 to avoid division by zero, respectively.}. \textcolor{black}{The $SSIM$ is in the range of [-1,1].} In particular, when $p$ is the same as $q$, $SSIM(p,q)$ is equal to 1. \textcolor{black}{Based on $SSIM$, the structural similarity loss is formulated as follows}:
\begin{equation}
  L_{SSIM} = \sum_{i=1}^{3} (1 - SSIM(f_{t}^{i}, f_{s}^{i}))\textcolor{black}{\,,}
  \label{eq:ssim-loss}
\end{equation}
\textcolor{black}{where $f_{t}^{i}$ and $f_{s}^{i}$ represent the $i\text{-}th$ stage feature maps of the teacher and the student, respectively.} Note that we only constrain $f_{t}^{i}$ and $f_{s}^{i}$ on normal regions.
Finally, we combine the above two losses to form the training objective of the teacher-student framework:
\begin{equation}
  L_{kd} = L_{cos} + L_{SSIM}\textcolor{black}{\,,}
  \label{eq:kd-loss}
\end{equation}

\subsubsection{The discrepancy map} \textcolor{black}{Compared with} existing knowledge distillation methods\cite{STPM, deng2022anomaly}, rather than directly taking the discrepancy map as the anomaly localization result, we serve it as \textcolor{black}{an additional robust cue for the segmentation task.} Taken a test image $I_m \in \mathbb{R}^{H \times W \times 3}$ as input, where $H$ and $W$ denote the height and width, respectively. \textcolor{black}{The teacher and the student} output corresponding feature representations $f_{t}^i$ and $f_{s}^i$. Then the discrepancy between $f_{t}^i$ and $f_{s}^i$ is measured by cosine and structural similarities. Specifically, the discrepancy map at location $(x,y)$ is \textcolor{black}{defined as}:
\begin{equation}
\label{Eq:distancemap}
\textcolor{black}{M^{i}_{(x,y)} =  2 - \frac{V_{t}^{i} \cdot V_{s}^{i}}{||V_{t}^{i}|| \cdot ||V_{s}^{i}||} -  SSIM(f_{t}^{i}, f_{s}^{i})_{(x,y)}\textcolor{black}{\,,}}
\end{equation}
where $SSIM(\cdot)_{(x,y)}$ represents the structural similarity of \textcolor{black}{the feature patches} centered at $(x,y)$, and the patch size is set to $11\times11$ in practice.
The discrepancy map of each layer $M^{i}$ is then upsampled to $H \times W$ and summed together as the final discrepancy map $\overline{M}\in \mathbb{R}^{H \times W}$.

Feeding the discrepancy map into the following segmentation head brings two desirable merits. On the one hand, the segmentation head can yield more discriminative representations under this strong guidance, thereby localizing anomalies more accurately.
On the other hand, the discrepancy map releases the segmentation head from being only constrained by \textcolor{black}{the synthetic appearance}, enhancing its perception of unseen anomalies. 
Such a strategy can greatly boost the robustness of our method under various simple and cheap synthesis strategies, making our method easy to use in practice. 

\begin{table*}[]
\caption{\textcolor{black}{Performance comparison on MVTecAD}}
\label{tab:compareSOTAmvtec}
\centering
\begin{tabular}{c|c|cccccccccc}
\toprule
 & Method & Backbone & Strategy & I-AUC & P-AUC & P-PRO & P-mAP & FLOPS & \textcolor{black}{FPS} & \#param. \\
                        \midrule
\multirow{4}{*}{\rotatebox{90}{KD}}    
& RDAD$^\ast$     & ResNet18 & \multirow{3}{*}{/} & 97.9 & 97.0            & 92.6   & 54.5  & 4.3G &  \textcolor{black}{127.0} & 15.9M    \\     
& STPM$^\dagger$      & ResNet18 &     & 95.0      & 96.1      & 86.8      & 47.4    & 3.7G  & \textcolor{black}{159.2} & 2.8M    \\
& MKDAD$^\ast$ & / & & 86.1 & 88.1 & 75.4 & 23.8 & 5.2G & \textcolor{black}{205.2} & \textbf{0.3M} \\
& STAD      & /        &    & 87.7          & 91.4       & /  & /        & 1948.1G & \textcolor{black}{4.71} & 26.4M    \\\hline
\multirow{8}{*}{\rotatebox{90}{\textcolor{black}{Self-Supervision}}} 
& CutPaste  & /        & $CutP$              & 96.1          & 96.0            & /             &     /   & /      &     \textcolor{black}{/}  & / &   \\
& DRÆM$^\ast$     & /   & $NSA_B$ & 84.1 & 95.1 & 84.0 & 45.9 & 198.4G  & \textcolor{black}{47.8} & 97.4M    \\
& NSA$^\ast$      & ResNet18 & $NSA_B$                & 95.6          & 96.3          & 90.5      & 58.7    & \textbf{2.5G} & \textcolor{black}{218.2}  & 11.5M    \\
& \textbf{DAF (Ours)}      & ResNet18 & $NSA_B$ & 97.1 & 97.5 & 92.4 & 66.0 & 6.8G & \textcolor{black}{93.6}  & 4.4M   \\\cline{2-12}
& DRÆM     & /        & $DRA$               & \textbf{98.0}   & 97.3 & /   &  68.4   & 198.4G  & \textcolor{black}{47.8} & 97.4M    \\
& NSA$^\ast$       & ResNet18 & $DRA$ & 92.0 & 93.6 & 86.7 & 58.0 & \textbf{2.5G}  & \textcolor{black}{218.2} & 11.5M \\
& TSDD      & /        & $DRA$               & 92.8          & 93.9          & /   & 60.7    & /   &  \textcolor{black}{/} & /    \\
& \textbf{DAF (Ours)}      & ResNet18 & $DRA$               & 97.6          & \textbf{98.1} & \textbf{93.0} & \textbf{68.5}  & 6.8G  & \textcolor{black}{93.6} & 4.4M   \\
                           \bottomrule
\end{tabular}
\end{table*}

\subsection{Segmentation head}

\textcolor{black}{The segmentation head aims at identifying the anomaly regions.} As illustrated in Fig.~\ref{fig:pipeline}(a), \textcolor{black}{the segmentation head} takes both the discrepancy map (\emph{i.e.}, $\overline{M}$) and \textcolor{black}{anomaly features} (\emph{i.e.}, $\{f_s^i\}_{i=1}^{3}$) as the input, where the discrepancy map indicates the location of anomalies while the anomaly features carry both low-level textual and high-level semantic information. 

The detailed fusion process \textcolor{black}{of} the discrepancy map and anomaly features is shown in Fig.~\ref{fig:pipeline}(b). 
The segmentation module processes them progressively in a coarse-to-fine manner via the three blocks, which is formulated as, 
\begin{equation}
  M_{S} = Seg_3(Seg_2(Seg_1(\overline{M} \textcircled{c} f_{s}^{3}) \textcircled{c} f_{s}^{2}) \textcircled{c} f_{s}^{1})\textcolor{black}{\,,}
  \label{eq:seg-df}
\end{equation}
where \textcolor{black}{$Seg_1,Seg_2,Seg_3$ denote the three blocks of the segmentation module,} \textcolor{black}{and} $\textcircled{c}$ denotes concatenation. 

\textcolor{black}{We apply the binary cross-entropy (BCE) loss as the segmentation loss.} Note that the anomaly (positive) pixels are much fewer than normal (negative) pixels. \textcolor{black}{To overcome} the imbalance of positive and negative pixels, a hard negative mining strategy is adopted. Mathematically, the segmentation loss $L_{seg}$ is defined as:
\begin{equation}
  L_{seg} = \sum_{i \in Sub} y_i \text{log} x_i + (1 - y_i) \text{log} (1 - x_i)\textcolor{black}{\,,}
  \label{eq:seg_loss}
\end{equation}
\textcolor{black}{where $Sub$ is a subset sampled from $M_{S}$}, $x_i$ is the predicted anomaly probability in $Sub$, and $y_i$ is its corresponding label.

\subsection{Auxiliary Supervision}
In the teacher-student framework, \textcolor{black}{the student is trained} to regress representations of \textcolor{black}{the teacher on normal regions}. However, there is no constraint for \textcolor{black}{the student} on anomalous regions, which hinders \textcolor{black}{the student} from providing discriminative representations for the segmentation head $Seg$.

To enhance the discrimination capability of \textcolor{black}{the student} on anomalous regions, we add \textcolor{black}{an auxiliary head after each student layer,} as shown in Fig.~\ref{fig:pipeline}(b).
\textcolor{black}{Each head $Aux_i$ takes its corresponding anomaly feature $f_s^i$ as input, and outputs the probability map} of the corresponding size. The training strategy is the same as \textcolor{black}{that of the segmentation head}, meaning that we adopt the BCE loss and hard mining strategy as the discriminative loss $L_{dis}$. 
Note that auxiliary heads are discarded during inference, incurring no extra cost. 

\subsection{Anomaly Localization}
During inference, we localize the anomalous regions from two aspects. 
Firstly, since the segmentation head $Seg$ is trained to \textcolor{black}{distinguish the normal distribution from that of the abnormal,} the segmentation probability map is expected to localize anomalies accurately.
Secondly, the discrepancy map between the teacher and the student can also localize anomalies, since the student is likely to demonstrate \textcolor{black}{inconsistency} with the teacher on anomalous patterns.
Since the discrepancy map is irrelevant to \textcolor{black}{the synthetic appearance}, combining it and the segmentation probability map desires both accurate and robust properties. 

In summary, the anomalies are located by incorporating the discrepancy map $\overline{M}$ with the segmentation probability map $M_{S}$. The final anomaly score map is formulated as:
\begin{equation}
\label{Eq:scoremap}
M_{Score} = G(\overline{M} + \lambda M_{S})\textcolor{black}{\,,}
\end{equation}
The $G$ means Gaussian smooth. $\lambda$ is set to 3 in practice.

\section{Experiments} \label{sec:experiments}
\subsection{\textcolor{black}{Experimental Setup}} 
\subsubsection{Datasets} MVTecAD~\cite{bergmann2019mvtec} contains 5,354 images, including ten object categories and five texture categories. The training set has 3,629 normal images, while the testing set contains 1,725 images\textcolor{black}{, covering both normal and anomaly images}. Pixel annotations are provided for the anomaly areas. 

DAGM~\cite{wieler2007weakly} includes ten categories of texture images. The training set contains anomaly images, and weak annotations are provided for anomalous regions. During training, we only use normal images. Since the annotations are coarse, we do not evaluate the localization performance on DAGM.

\begin{table*}[]
\centering
\caption{\textcolor{black}{Localization performance comparison under simple and cheap synthesis strategies}}
\label{tab:overfitOnMVTEC}
\setlength{\tabcolsep}{1.45mm}
\begin{tabular}{c|c|l|ccccccccccccccc|c}
\toprule
& Metric  & Method & Carp. & Grid & Leath. & Tile & Wood & Bottle & Cable & Caps. & Haze. & Metal. & Pill & Screw & Tooth. & Trans. & Zip. & Mean \\
\midrule
\multirow{9}{*}{\rotatebox{90}{\emph{Simple Texture}}} & \multirow{3}{*}{P-AUC} & DRÆM  & /  & / & / & / & / & / & / & / & / & / & / & / & / & /  & /  & 92.6 \\
&   & NSA$^\ast$    & 91.3   & 80.3 & 90.6    & 95.0   & 83.2 & 92.9   & 76.4  & 83.8    & 91.4     & 96.7     & 80.9 & 73.5  & 96.8       & 68.6       & 64.5   & 84.4 \\
&  & DAF  & \textbf{98.8}   & \textbf{98.8} & \textbf{99.6}    & \textbf{97.7} & \textbf{96.8} & \textbf{98.7}   & \textbf{97.2}  & \textbf{97.4}    & \textbf{99.0}     & \textbf{99.1}     & \textbf{98.0} & \textbf{99.0}  & \textbf{99.1}       & \textbf{89.8}       & \textbf{98.3}   & \textbf{97.8} \\\cline{2-19} 
& \multirow{3}{*}{P-PRO} & DRÆM  & /   & / & /  & / & / & /  & /  & /   & /    & /     & / & /  & /   & /       & /  & / \\
&                      & NSA$^\ast$    & 79.9   & 54.2 & 88.7    & \textbf{93.9} & 79.8 & 78.1   & 62.9  & 66.8    & 89.5     & 91.1     & 78.7 & 38.1  & 75.3       & 53.7       & 36.4   & 71.1 \\
&                      & DAF  & \textbf{94.2}   & \textbf{94.2} & \textbf{98.5}    & 92.7 & \textbf{92.9} & \textbf{94.5}   & \textbf{89.9}  & \textbf{84.3}    & \textbf{95.6}     & 93.9     & \textbf{92.4} & \textbf{95.1}  & 90.3       & \textbf{78.7}       & \textbf{91.7}   & \textbf{91.9} \\\cline{2-19} 
& \multirow{3}{*}{P-mAP} & DRÆM  & /   & / & /  & / & / & /  & /  & /   & /    & /     & / & /  & /   & /       & /  & 56.5 \\
&                      & NSA$^\ast$    & 47.7 & 8.2 & 42.4  & \textbf{89.1} & 68.0 & 62.3  & 16.8  & 30.1   & 63.9    & 91.5     & 28.3 & 2.2  & 55.7   & 22.9       & 13.8  & 42.9 \\
&                      & DAF & \textbf{67.1} & \textbf{43.1} & \textbf{66.8}  & 84.3 & \textbf{73.2} & \textbf{81.6}  & \textbf{55.8}  & \textbf{41.0}   & \textbf{67.2}    & \textbf{93.8}     & \textbf{54.1} & \textbf{34.2}  & \textbf{66.0}   & \textbf{55.3}       & \textbf{49.7}  & \textbf{62.2} \\\hline
\multirow{9}{*}{\rotatebox{90}{\emph{Simple Shape}}}  & \multirow{3}{*}{P-AUC} & DRÆM$^\ast$  & 79.8   & 97.6 & 91.5    & 80.2 & 74.6 & 74.5   & 71.3  & 63.4    & 81.3     & 73.0     & 72.2 & 81.7  & 87.5       & 71.1       & 82.4   & 78.8 \\
                       &                      & NSA$^\ast$    & 76.6   & 95.4 & 95.3    & 77.4 & 78.0 & 73.6   & 61.4  & 63.4    & 87.2     & 59.3     & 67.6 & 82.8  & 79.5       & 67.4       & 75.8   & 76.0 \\
                       &                      & DAF  & \textbf{98.6}   & \textbf{99.3} & \textbf{99.7}    & \textbf{98.6} & \textbf{96.4} & \textbf{98.5}   & \textbf{96.7}  & \textbf{95.7}    & \textbf{98.8}     & \textbf{98.5}     & \textbf{97.5} & \textbf{98.8}  & \textbf{99.0}       & \textbf{88.1}       & \textbf{99.1}   & \textbf{97.6} \\\cline{2-19} 
                       & \multirow{3}{*}{P-PRO} & DRÆM$^\ast$  & 70.8   & 94.9 & 89.4    & 59.4 & 71.0 & 61.3   & 30.6  & 55.6    & 78.1     & 56.3     & 75.8 & 66.9  & 78.9       & 53.1       & 70.9   & 67.5 \\
                       &                      & NSA$^\ast$    & 65.1   & 88.0   & 93.2    & 53.3 & 78.0 & 61.1   & 38.5  & 60.8    & 87.9     & 25.5     & 53.0 & 64.5  & 71.6       & 49.7       & 59.9   & 63.3 \\
                       &                      & DAF  & \textbf{96.1}   & \textbf{97.3} & \textbf{99.0}    & \textbf{95.0} & \textbf{93.1} & \textbf{95.1}   & \textbf{90.3}  & \textbf{81.7}    & \textbf{96.5}     & \textbf{96.0}     & \textbf{93.3} & \textbf{94.2}  & \textbf{93.4}       & \textbf{75.2}       & \textbf{96.0}   & \textbf{92.8} \\\cline{2-19} 
& \multirow{3}{*}{P-mAP} & DRÆM$^\ast$  & 23.5 & 49.1 & 55.5  & 29.4 & 32.5 & 36.8  & 5.8  & 6.3   & 31.2   & 22.1   &  18.8 & 6.9  & 25.5  & 11.3       & 33.3  & 25.9 \\
&                      & NSA$^\ast$    & 12.8 & 45.3 & 34.2  & 19.1 & 53.8 & 43.4  & 4.7  & 13.3   & 42.5    & 13.7     & 6.2 & 9.6  & 16.9   & 12.3       & 30.1  & 23.9 \\
&                      & DAF & \textbf{70.1}   & \textbf{57.4} & \textbf{69.8}  & \textbf{88.2} & \textbf{65.9} & \textbf{80.1}  & \textbf{45.5}  & \textbf{24.1}   & \textbf{60.9}   & \textbf{86.8}     & \textbf{72.1} & \textbf{42.1}  & \textbf{54.5}   & \textbf{51.7}   & \textbf{71.1}  & \textbf{62.7} \\\hline
\multirow{9}{*}{\rotatebox{90}{\emph{Simple Texture-Shape}}}  & \multirow{3}{*}{P-AUC} & DRÆM$^\ast$  &  58.6  & 95.0 &  71.6   & 85.4 & 69.2 & 80.2   & 47.2 &  58.4   &  55.1   &  66.5   & 46.9 &  80.8 &  59.8  &  57.9  &  59.9 & 66.2 \\
&  & NSA$^\ast$    &  59.7 & 60.6 &  67.5   &  55.7 & 58.4 &  51.4  & 52.2 & 50.0    &   59.8  &  52.7   &  48.0 & 58.1  &   53.5    &  47.8   &  60.2  & 55.7 \\
&     & DAF  & \textbf{99.0}   &  \textbf{98.9} &  \textbf{99.6}   &  \textbf{98.0} & \textbf{96.2} & \textbf{98.4}   &  \textbf{96.2} &   \textbf{97.6}  &  \textbf{98.6}   &  \textbf{98.7}   &  \textbf{97.5} &  \textbf{99.1} &   \textbf{98.9}    &  \textbf{88.2}   & \textbf{98.6}  & \textbf{97.6} \\\cline{2-19} 
& \multirow{3}{*}{P-PRO} & DRÆM$^\ast$  &  40.0 & 90.3 &   69.2  & 58.2  & 44.2 &  57.6  & 30.6 &  43.4   &  38.4   &  60.1   & 53.9 &  64.3 &   70.7    &  43.7   & 38.9  & 53.6 \\
&  & NSA$^\ast$    &  35.2 & 32.4 &  56.7  & 36.0 & 53.6 &  24.7  & 18.6 &  23.9   &  23.5   &   22.6  & 32.5 & 25.0  &   20.9    & 22.5    &  29.3  & 30.5 \\
&      & DAF   &  \textbf{96.1}  & \textbf{95.2} & \textbf{98.8}   & \textbf{93.4} & \textbf{93.1} & \textbf{94.5}   & \textbf{88.9} & \textbf{87.6}    &  \textbf{94.1}   &   \textbf{92.8}  & \textbf{94.4} & \textbf{95.5}  &  \textbf{91.9}   &  \textbf{75.3}  &  \textbf{92.9} & \textbf{92.3} \\\cline{2-19} 
& \multirow{3}{*}{P-mAP} & DRÆM$^\ast$  & 6.5 & \textbf{47.1} & 42.7  & 46.4 & 15.6 & 42.0  & 4.2  & 8.3   & 19.9    & 31.2  & 14.3 & 13.8  & 25.2   & 10.8      & 17.3 & 23.0 \\
&                      & NSA$^\ast$    & 2.9 & 3.2 & 8.2  & 14.3 & 16.8 & 7.2  & 3.0  & 1.0   & 2.7   & 12.7     & 3.3 & 0.4  & 1.9   & 5.0     & 3.9  & 5.8 \\
&                      & DAF & \textbf{71.6} & 42.5 & \textbf{67.3} & \textbf{85.1} & \textbf{63.1} & \textbf{81.6}  & \textbf{44.7}  & \textbf{35.7}   & \textbf{60.2}    & \textbf{88.4}    & \textbf{72.7} & \textbf{29.9}  & \textbf{53.3}   & \textbf{53.7}      & \textbf{53.0}   & \textbf{60.2} \\
\bottomrule
\end{tabular}
\end{table*}

\subsubsection{Model Training}
All the images are resized to $256 \times 256$. The weights of the teacher are frozen, and the remaining components are trained using AdamW for 1,200 epochs. The batch size is set to 8, and the weight decay is set to $1e{-5}$. The learning rate is gradually increased to $2e{-4}$ in 50 epochs and multiplied by 0.2 after 700 and 1,000 epochs. We first follow \textcolor{black}{the} synthetic strategies in DRÆM~\cite{Zavrtanik_2021_ICCV} and NSA~\cite{Schlter2022NaturalSA} \textcolor{black}{and then} adopt a series of simple synthetic approaches to further investigate the effectiveness of our method. \textcolor{black}{Note that DRÆM~\cite{Zavrtanik_2021_ICCV} introduces an external dataset (\emph{i.e.} DTD~\cite{DTDdata}) for synthesizing.}

\subsubsection{Evaluation Metrics}
Following Salehi \etal~\cite{multiresolution}, we evaluate the performance of anomaly detection and localization by image-level AUC (I-AUC) and pixel-level AUC (P-AUC), respectively.
Meanwhile, following ~\cite{bergmann2019mvtec} and ~\cite{Zavrtanik_2021_ICCV}, we also focus on the Per-Region-Overlap (P-PRO) and mean Average Precision (P-mAP) to further evaluate localization accuracy.
\subsubsection{Baselines}
\textcolor{black}{We mainly compare DAF with knowledge distillation-based methods and self-supervision-based methods. For ease of description, we abbreviate these two methods in the following tables as KD and Self-Sup, respectively.}

\textcolor{black}{(1) Knowledge distillation-based methods:} STAD\cite{bergmann2020uninformed}, STPM\cite{STPM}, MKDAD\cite{multiresolution}, RDAD\cite{deng2022anomaly}. \textcolor{black}{These methods} compare activations of the teacher and student, where features of anomaly regions are distinct.

\textcolor{black}{(2) Self-supervision-based methods:} CutPaste\cite{cutpaste}, DRÆM\cite{Zavrtanik_2021_ICCV}, NSA\cite{Schlter2022NaturalSA}. \textcolor{black}{These methods} introduce extra supervision through synthetic anomalies to help proxy tasks such as segmentation.

\begin{table*}
\centering
\caption{Detection performance on DAGM~\cite{wieler2007weakly} under different synthesis strategies}
\label{tab:dagm-img}
\begin{tabular}{c|ccccccccccccc}
\toprule
 & Method & \textcolor{black}{Strategy}             & Class1        & Class2        & Class3        & Class4        & Class5        & Class6        & Class7       & Class8        & Class9        & Class10       & Mean           \\
\midrule
\multirow{2}{*}{\rotatebox{90}{KD}} & RDAD$^\ast$            & /                               & 95.3          & 99.2          & 80.6          & 100.0           & 78.1          & 91.6          & 99.5         & 63.1          & 95.3          & 99.1          & 90.2          \\
& STPM$^\dagger$           & /                               & 86.1          & 98.7          & 88.7          & 100.0           & 82.3          & 94.4          & 99.8         & 65.6          & 91.7          & 99.7          & 90.7          \\\hline
\multirow{12}{*}{\textcolor{black}{\rotatebox{90}{Self-Supervision}}} & DRÆM            & \multirow{3}{*}{$DRA$}            & /             & /             & /             & /             & /             & /             & /            & /             & /             & /             & 99.0          \\
& NSA$^\ast$       &     &  90.6  & 99.2  &      99.8         &       \textbf{100.0}        &        \textbf{98.1}       & 99.9    &   \textbf{100.0}      &   \textbf{99.5}            &      51.4        &     99.7        &      93.8     \\
& DAF  &    &  \textbf{99.5}    &  \textbf{99.8}  &  \textbf{100.0}    &     \textbf{100.0}   &    96.7  &   \textbf{100.0}    &  \textbf{100.0}  &  97.1   &    \textbf{98.9}    &   \textbf{99.9}   &   \textbf{99.2}  \\\cline{2-14}
& DRÆM$^\ast$            & \multirow{3}{*}{$NSA_B$}            & 45.1          & \textbf{100.0}  & 88.5          & 80.0            & 84.1          & \textbf{100.0}  & \textbf{100.0}          & 83.9          & 91.2          & 87.9          & 86.1          \\
& NSA$^\ast$            &                                 & 52.6          & 99.3          & 85.1          & 97.8          & 86.7          & \textbf{100.0}  & 57.4         & 83.2          & 62.1          & 93.8          & 81.8          \\
& DAF           &                                 & \textbf{85.7} & \textbf{100.0}  & \textbf{99.6} & \textbf{99.8} & \textbf{93.8} & \textbf{100.0}  & \textbf{100.0} & \textbf{100.0}  & \textbf{99.7} & \textbf{99.9} & \textbf{97.9} \\\cline{2-14}
& DRÆM$^\ast$            & \multirow{3}{*}{\emph{Simple Texture}} & 54.9          & 45.6          & 98.8          & 68.8          & \textbf{99.3} & 57.0            & 93.4         & 87.4          & 53.4          & 72.6          & 73.1          \\
& NSA$^\ast$            &                                 & 50.6          & 50.8          & 80.5          & 51.4          & 72.7          & 64.6          & \textbf{100.0} & \textbf{97.1} & 53.3          & 76.1          & 69.7          \\
& DAF           &                                 & \textbf{97.6} & \textbf{97.1} & \textbf{99.7} & \textbf{100.0}  & 76.1          & \textbf{96.7} & \textbf{100.0} & 88            & \textbf{91.8} & \textbf{99.7} & \textbf{94.7} \\\cline{2-14}
& DRÆM$^\ast$            & \multirow{3}{*}{\emph{Simple Shape}}   & \textbf{99.4} & 90.2          & 89.1          & 96.7          & \textbf{100.0}  & 86.9          & \textbf{100.0} & 92.3          & 48.2          & 77.7          & 88.1          \\
& NSA$^\ast$             &                                 & 52.2          & 44.2          & 36.6          & 29.6          & 28.5          & 58.8          & 20.5         & 46.5          & 52.8          & 25.1          & 39.5          \\
& DAF           &                                 & 93.6          & \textbf{100.0}  & \textbf{100.0}  & \textbf{100.0}  & 97.8          & \textbf{100.0}  & \textbf{100.0} & \textbf{97.0} & \textbf{94.8} & \textbf{99.9} & \textbf{98.3} \\
\bottomrule
\end{tabular}
\end{table*}

\subsection{Evaluation on MVTecAD using complicated strategies}
\subsubsection{Quantitative analysis} Table~\ref{tab:compareSOTAmvtec} reports the detection and localization performance \textcolor{black}{on the MVTecAD~\cite{bergmann2019mvtec} dataset}\footnote{\textcolor{black}{$\ast$ represents the reproduced result using the official code. $\dagger$ denotes the reproduced results using the unofficial code.}}. In our method, the image-level score is acquired by the mean of the top 50 values of \textcolor{black}{the anomaly score map $M_{Score}$.} 
\begin{figure}[!t]
    \centering
    \includegraphics[width=1.0\columnwidth]{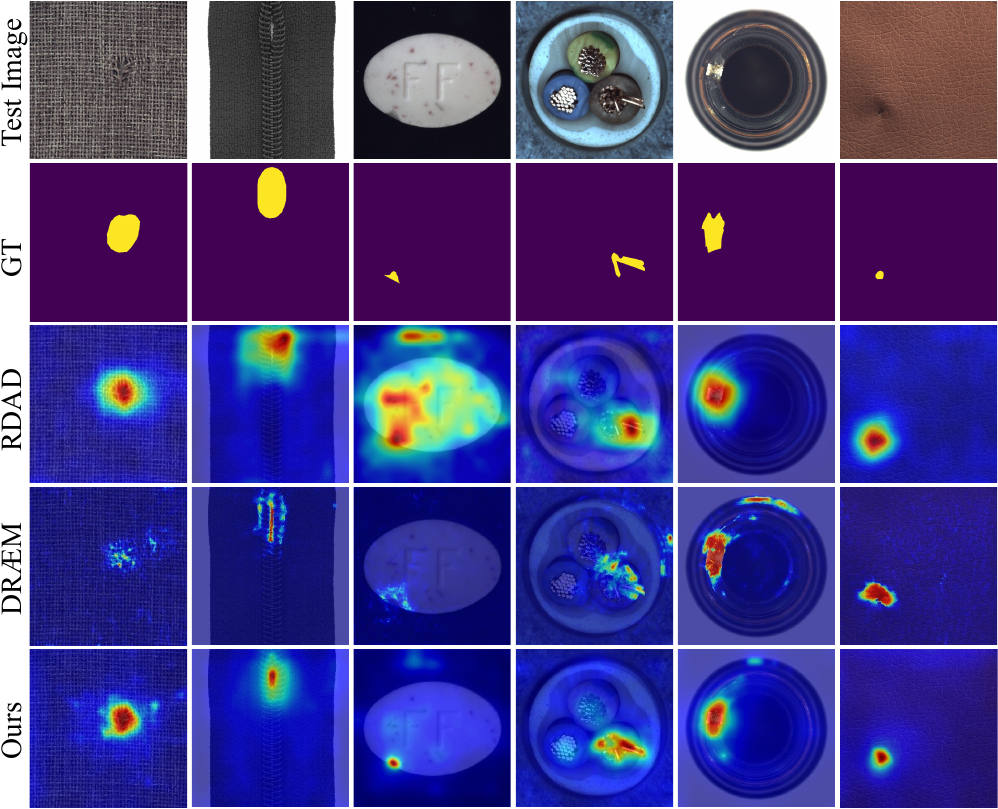}
    \caption{\textcolor{black}{Visualization of localization results.}}
\label{figure:visual-custom}
\end{figure}
\textcolor{black}{Compared with} knowledge distillation methods, our method achieves a comparable result of 97.6\% I-AUC with RDAD~\cite{deng2022anomaly} on the detection task. In terms of the localization task, benefiting from \textcolor{black}{the discriminative capacity} of the segmentation head $Seg$, our approach surpasses RDAD by 1.1\% P-AUC and 14.0\% P-mAP. Compared with self-supervised methods, while using \textcolor{black}{the synthesis strategy of $NSA_B$}, we exceed NSA~\cite{Schlter2022NaturalSA} by 1.5\% I-AUC and 7.3\% P-mAP, respectively. Furthermore, compared with DRÆM~\cite{Zavrtanik_2021_ICCV}, \textcolor{black}{our method achieves significant improvements} of 13.0\% I-AUC and 20.1\% P-mAP. 
Notably, when using the synthesis strategy of $DRA$, our detection performance is slightly lower than DRÆM~\cite{Zavrtanik_2021_ICCV} by 0.4\%. \textcolor{black}{However, our localization performance exceeds it by 0.8\% P-AUC.}
Please note that the computation complexity (FLOPS), the model size (\#param), and the FPS of DRÆM are much higher than ours.
\textcolor{black}{When training with} synthetic data, the supervised method TSDD~\cite{bovzivc2021end} obtains relatively poor performance, indicating that the distribution gap with \textcolor{black}{real anomalies may hurt its generalization.}

\subsubsection{Qualitative analysis} Fig.~\ref{figure:visual-custom} shows localization results of the existing state-of-the-art knowledge distillation method RDAD~\cite{deng2022anomaly}, \textcolor{black}{the self-supervised method DRÆM~\cite{Zavrtanik_2021_ICCV} and our method.} \textcolor{black}{Both DRÆM and our method are trained using the $DRA$ strategy.} It can be observed that RDAD can roughly localize anomalies, \textcolor{black}{but the localization areas} tend to be larger than the ground truth (Row 3, Columns 2,4,5,6). 
DRÆM can sometimes deliver clear decisions for normal and abnormal areas (Row 4, Column 5), but it also fails in some cases (Row 4, Columns 1-4).
The last row suggests that our method achieves the most accurate localization, performing well across various anomalies and closely matching the ground truth.

\begin{table}
\centering
\caption{Ablation study on the effectiveness of each component}
\label{ablation:arch.}
\begin{tabular}{ccccc}
\toprule
Components & I-AUC & P-AUC      & P-PRO        & P-mAP   \\
\midrule
\emph{only $T$-$S$}                 & 93.6    & 96.3     & 86.1      & 48.6   \\
\emph{only $Seg$}        & 95.1    & 95.2       & 75.1    & 62.5    \\
\emph{W/O $Aux$}   & \textbf{97.7}    & 98.0            & 92.6    & 66.6   \\
DAF    & 97.6    & \textbf{98.1} & \textbf{93.0} & \textbf{68.5} \\
\bottomrule
\end{tabular}
\end{table}

\begin{table}[]
\centering
\caption{\textcolor{black}{Ablation study on the discrepancy map}}
\label{ablation:discrepancyrobust}
\begin{tabular}{ccccccc}
\toprule
\textcolor{black}{Strategy} & $\overline{M}$  &  $M_S$ & I-AUC & P-AUC & P-PRO & P-mAP \\
\midrule
\multirow{3}{*}{$DRA$} & \checkmark & &  95.4    & 97.4      & 90.7       & 52.8     \\
  &  & \checkmark & 96.8    &  95.5        & 78.4     & 67.1       \\ 
 & \checkmark & \checkmark   & \textbf{97.6}      & \textbf{98.1}      & \textbf{93.0}        & \textbf{68.5}    \\\hline
\multirow{3}{*}{\emph{Simple Texture}} & \checkmark & &  97.4  &  97.4 &  90.8  &  52.1 \\
 &  & \checkmark &  95.3   &  89.9  &   55.6     &  52.9 \\ 
 & \checkmark & \checkmark &  \textbf{97.5}   &  \textbf{97.8}  & \textbf{91.9} & \textbf{62.2} \\\hline
\multirow{3}{*}{\emph{Simple Shape}} & \checkmark & &  92.9  &   97.0   &   89.9    &  48.7 \\
 &  & \checkmark &   94.0  &  80.8    &  72.2    &  40.0 \\ 
 & \checkmark & \checkmark &   \textbf{97.0}  &  \textbf{97.6} &  \textbf{92.8} & \textbf{62.7} \\
\bottomrule
\end{tabular}
\end{table}

\subsection{Evaluation on MVTecAD using simple strategies}
\emph{Simple Texture} means that \textcolor{black}{the textures of the synthetic anomalies are random colors, rather than} natural textures sampled from the DTD dataset~\cite{DTDdata}.
\textcolor{black}{\emph{Simple Shape} indicates that the shapes of synthetic anomalies are rectangles, in contrast to irregular shapes.}
\emph{Simple Texture-Shape} replaces shapes and textures with rectangles and randomly sampled colors.

\subsubsection{Quantitative analysis} 

\textcolor{black}{\textcolor{black}{Table}~\ref{tab:overfitOnMVTEC} reports the results. Under the \emph{Simple Texture} strategy, it can be observed that our method achieves competitive results of 97.8\% P-AUC, 91.9\% P-PRO and 62.2\% P-mAP, while DRÆM~\cite{Zavrtanik_2021_ICCV} and NSA~\cite{Schlter2022NaturalSA} perform relatively lower. Specifically, \textcolor{black}{compared with DRÆM}, we bring 5.2\% (97.8\% \emph{vs.} 92.6\%) P-AUC and 5.7\% (62.2\% \emph{vs}. 56.5\%) P-mAP gains, and outperform NSA by 13.4\% P-AUC and 19.3\% P-mAP. Similarly, \textcolor{black}{compared with} these two prior arts, we also achieve significant improvements under the \emph{Simple Texture} and \emph{Simple Texture-Shape} strategies. These results demonstrate the superior robustness of our framework.}

\subsubsection{Discussions} NSA~\cite{Schlter2022NaturalSA} carefully designs a synthesis strategy to simulate natural images. Since the optimization process is only influenced by the anomalous appearance, the model is likely to \textcolor{black}{overfit to} the synthetic anomalies. Hence, it can perform well if the real anomalies are similar to the synthetic ones. However, it is challenging to localize anomalies when \textcolor{black}{there is a distribution discrepancy between them.}
DRÆM~\cite{Zavrtanik_2021_ICCV} attempts to solve the \textcolor{black}{overfitting} issue by jointly using \textcolor{black}{the synthetic anomaly data} and anomaly-free reconstruction. However, the reconstruction model also suffers from the \textcolor{black}{overfitting} problem, failing to restore anomalies when unseen anomalies occur.
Our method benefits from the discrepancy map, which inherently indicates the location of anomaly regions, providing strong evidence for the segmentation head. \textcolor{black}{This alleviates} its dependence on \textcolor{black}{the} synthetic appearance. Moreover, the supplement of the discrepancy map can further promote robustness.

\begin{table}
    \centering
    \caption{Impact of the discrepancy map $\overline{M}$ on segmentation}
    \label{ablation:discrep}
    
    \begin{tabular}{cccccc}
        \toprule
        \textcolor{black}{Strategy}  &  $\overline{M}$  & I-AUC   & P-AUC     & P-PRO     & P-mAP     \\
        \midrule
         \multirow{2}{*}{$DRA$}  &            & \textbf{97.6} & 98.0            & 92.9          & 67.3          \\
              &                  \checkmark                 & \textbf{97.6} & \textbf{98.1} & \textbf{93.0}   & \textbf{68.5} \\\hline
                 \multirow{2}{*}{\emph{Simple Texture}} & &  96.5          & 97.7          & 91.3          & 60.1          \\
                &                   \checkmark              & \textbf{97.5} & \textbf{97.8} & \textbf{91.9} & \textbf{62.2} \\\hline
                 \multirow{2}{*}{\emph{Simple Shape}}   & &  96.5          & 97.5          & 92.6          & 62.2          \\
               &                  \checkmark                & \textbf{97.0}   & \textbf{97.6} & \textbf{92.8} & \textbf{62.7} \\
        \bottomrule
    \end{tabular}
\end{table}

\subsection{Evaluation on DAGM}
\textcolor{black}{Table}~\ref{tab:dagm-img} shows the detection performance on DAGM\textcolor{black}{~\cite{wieler2007weakly}}. The two knowledge distillation methods RDAD~\cite{deng2022anomaly} and STPM~\cite{STPM} achieve \textcolor{black}{comparable performance} (90.2\% \emph{vs.} 90.7\%). 
\textcolor{black}{When training with $DRA$}, the self-supervised methods DRÆM~\cite{Zavrtanik_2021_ICCV} and NSA~\cite{Schlter2022NaturalSA} outperform knowledge distillation methods by a large margin.
Our method obtains \textcolor{black}{the state-of-the-art} performance of 99.2\%. We assume that $DRA$ can synthesize anomalies that match real ones during training, helping \textcolor{black}{the self-supervised methods} to yield clear decision boundaries to classify the normal and abnormal regions. 
However, these methods demonstrate unsatisfactory results when the synthetic strategies are simpler, \emph{e.g.}, \emph{Simple Texture} and \emph{Simple Shape}. In contrast, our method demonstrates strong robustness to these synthesis strategies. 
The improvement is thanks to the discrepancy map, which reveals the location of anomalies, releasing the segmentation task from the \textcolor{black}{constraint of the} anomaly appearance. Additionally, the supplement of the discrepancy map during inference has further enhanced robustness.

\subsection{Ablation Study}
For simplicity and fairness, we conduct all the ablation studies on MVTecAD~\cite{bergmann2019mvtec}. %

\subsubsection{Effectiveness of different components}
\textcolor{black}{Table}~\ref{ablation:arch.} displays the impacts of different components. \textcolor{black}{\textit{Only} $T\text{-}S$ indicates that only the teacher-student framework remains, and only the discrepancy map is employed for evaluation.} 
\textcolor{black}{We also only keep the student and the segmentation head to build a segmentation network (\emph{Only $Seg$}).
To demonstrate the effectiveness of the auxiliary heads, we remove them from the pipeline (\emph{W/O $Aux$}).}
Compared with \emph{only $Seg$}, \emph{only $T\text{-}S$} shows its advantage in P-AUC and P-PRO. In contrast, \emph{only $Seg$} brings higher P-mAP. 
\textcolor{black}{\emph{W/O $Aux$} brings gains in both detection and localization.} \textcolor{black}{By incorporating the auxiliary heads during training (Ours), we \textcolor{black}{achieve the best localization performance.} This observation indicates that the auxiliary heads promote the student to yield more discriminative representations for anomalies.}

\begin{table}
    \centering
    \caption{\textcolor{black}{Ablation study on the hyperparameter under $DRA$ strategy}}
    \label{ablation:aug-param}
    \begin{tabular}{cccccccc}
        \toprule
    Method  & DTD & Perlin  & $\beta$  & I-AUC & P-AUC & P-PRO & P-mAP \\
        \midrule
    \multirow{2}{*}{DRÆM} &  \checkmark   & \checkmark       & \checkmark                    & 98.0      & 97.3      & /         & 68.4      \\
                & \checkmark   & \checkmark       &                      & 97.4    & 95.0        & /         & 64.5      \\\cline{1-8}
    \multirow{2}{*}{DAF}  & \checkmark   & \checkmark       & \checkmark                    & 97.6    & 98.1      & 93.0        & 68.5      \\
        & \checkmark   & \checkmark       &                      & 97.6    & 97.8      & 92.4      & 64.5        \\
       \bottomrule
    \end{tabular}
\end{table}

\begin{table}
    \centering
    \caption{Comparison with the model \textcolor{black}{ensemble}}
    \label{ablation:ensemble}
    \begin{tabular}{cccccc}
        \toprule
        \textcolor{black}{Strategy} & Method & I-AUC & P-AUC & P-PRO  & P-mAP  \\
        \midrule
        \multirow{2}{*}{$DRA$}      & \emph{Ensem.} &  96.8    & 97.7     & 91.7 & 65.1 \\
             &         DAF           & \textbf{97.6}    & \textbf{98.1}      & \textbf{93.0}   & \textbf{68.5} \\\hline
        \multirow{2}{*}{\emph{Simple Texture}} & \emph{Ensem.} &  95.8    & 97.4      & 90.1 & 58.4 \\
              & DAF    & \textbf{97.5}    & \textbf{97.8}      & \textbf{91.9} & \textbf{62.2} \\\hline
         \multirow{2}{*}{\emph{Simple Shape}}   & \emph{Ensem.} & 93.5      & 97.5      & 91.7 & 61.2 \\
            & DAF    & \textbf{97.0}      & \textbf{97.6}      & \textbf{92.8} & \textbf{62.7} \\
        \bottomrule
    \end{tabular}
\end{table}

\subsubsection{Effectiveness of the discrepancy map}
 We further conduct several settings to prove that the discrepancy map $\overline{M}$ is a useful supplement to the segmentation probability map $M_S$. \textcolor{black}{Table}~\ref{ablation:discrepancyrobust} shows that with $\overline{M}$, we have observed 2.6\% P-AUC, 14.6\% P-PRO and 1.4\% P-mAP improvements under the $DRA$ strategy. It can also be observed that the combination of \textcolor{black}{the discrepancy map $\overline{M}$ and the segmentation probability map $M_S$} brings significant increases in both detection and \textcolor{black}{localization performance} under the simple strategies. 

\subsubsection{Influence of the discrepancy map on segmentation}
To investigate the influence of the discrepancy map on the segmentation task, we remove it from the input of the segmentation head $Seg$. As shown in \textcolor{black}{Table}~\ref{ablation:discrep}, the performance suffers from degradation \textcolor{black}{when the discrepancy map is not incorporated}. For instance, under the \emph{Simple Texture} strategy, the detection performance drops from 97.5\% to 96.5\%, while the localization P-mAP reduces by 2.1\%. 
We hypothesize that our improvements lie in the discrepancy map, which provides \textcolor{black}{an effective cue} for the segmentation task. \textcolor{black}{This alleviates} the segmentation head from the limitation of \textcolor{black}{the} synthetic appearance, \textcolor{black}{thereby enhancing its capacity to identify previously unseen anomalies.}

\subsubsection{Influence of the hyperparameter in the synthesis strategy}
Following DRÆM~\cite{Zavrtanik_2021_ICCV}, we evaluate how the hyperparameter $\beta$ impacts the performance of our method. The result is reported in Table~\ref{ablation:aug-param}, \textcolor{black}{where DTD and Perlin noise are utilized to simulate the appearance and shape of anomalies in the $DRA$ strategy.} $\beta$ controls the opacity in blending.
Compared with DRÆM, our method is more robust on $\beta$. We keep the same P-mAP as DRÆM, but show advantages in I-AUC (97.6\% \emph{vs.} 97.4\%) and P-AUC (97.8\% \emph{vs.} 95.0\%) \textcolor{black}{when training} via strategy $DRA$ without $\beta$. 

\subsubsection{Model Ensemble \emph{vs.} End-to-End} \textcolor{black}{Table}~\ref{ablation:ensemble} reports the performance of the model ensemble (\emph{Ensem.}). Here we train the teacher-student ($T\text{-}S$), and a segmentation model with auxiliary heads alone, then add the discrepancy and \textcolor{black}{the segmentation probability map} for evaluation. 
Compared with \emph{Ensem.}, we achieve improvements in both anomaly detection and localization. 
\textcolor{black}{The result suggests that with the guidance of the discrepancy map, our end-to-end framework could distinguish normal and abnormal regions more accurately.}

\begin{table}[]
\centering
\caption{Impact of initialization for the student}
\label{ablation:ini}
\begin{tabular}{cccccc}
\toprule
 \textcolor{black}{Strategy}      & Initial Param. & I-AUC     & P-AUC     & P-PRO     & P-mAP     \\
\midrule
 \multirow{2}{*}{$DRA$}  & Pretrained   & \textbf{97.6} & 97.9          & 91.9          & 64.3          \\
  &        Random              & \textbf{97.6} & \textbf{98.1} & \textbf{93.0}   & \textbf{68.5} \\\hline
\multirow{2}{*}{\emph{Simple Texture}} & Pretrained   &  96.5          & 97.2          & 89.5          & 54.2          \\
       &           Random     & \textbf{97.5} & \textbf{97.8} & \textbf{91.9} & \textbf{62.2} \\\hline
\multirow{2}{*}{\emph{Simple Shape}}    & Pretrained   &  96.9      & 97.3   & 92.0   & 58.8       \\
       &               Random      & \textbf{97.0}   & \textbf{97.6} & \textbf{92.8} & \textbf{62.7} \\
\bottomrule
\end{tabular}
\end{table}

\begin{table}[]
\centering
\setlength\tabcolsep{10pt}
\caption{Ablation study on the \textcolor{black}{losses} in the T-S framework}
\label{ablation:loss-func}
\resizebox{0.95\linewidth}{!}{
\begin{tabular}{ccccc}
\toprule
Loss   & I-AUC & P-AUC & P-PRO & P-mAP \\
\midrule
$L_{Cos}$ & 97.3      & 97.5      & 91.1      & 63.4     \\
$L_{SSIM}$   & 97.4      & 98.0        & 91.5      & 67.0       \\
DAF  & \textbf{97.6}      & \textbf{98.1}      & \textbf{93.0}        & \textbf{68.5}    \\
\bottomrule
\end{tabular}}
\end{table}

\subsubsection{Influence of initialization for the student}
\textcolor{black}{Table}~\ref{ablation:ini} studies the impact of parameter initialization on the student $S$. As shown, the student initialized randomly performs better, indicating that pre-trained parameters hinder $S$ from yielding disparate representations from the teacher on anomalies. We assume that the discrepancy map derived from $T\text{-}S$ provides only \textcolor{black}{limited cue} for the segmentation head $Seg$ in this case, resulting in reduced robustness of $Seg$.

\subsubsection{Influence of the distillation loss}
\textcolor{black}{Table}~\ref{ablation:loss-func} shows the effectiveness of the structural similarity ($SSIM$) loss.
Since the cosine similarity loss solely distills knowledge at each position in isolation, we introduce $SSIM$ loss to consider neighboring vectors. The study reveals the effectiveness of \textcolor{black}{the} $SSIM$ loss. \textcolor{black}{DAF} achieves the best results when the cosine similarity and $SSIM$ are adopted simultaneously.

\subsubsection{Limitations} While synthesized anomalies boost performance, the training time increases due to \textcolor{black}{the} computationally expensive synthesizing steps. Although adopting simple and cheap ones accelerates the process, it is still slower than unsupervised methods that are solely trained on normal images.

\subsubsection{Future work} \textcolor{black}{In future work, investigating lightweight \textcolor{black}{architectures} and developing real-time approaches will be important for industrial applications. Moreover, integrating the rich knowledge derived from large models such as CLIP~\cite{radford2021learning} into our method will be helpful for further enhancing the generalization capacity.}

\section{Conclusion}\label{sec:conclusion}
In this paper, we have observed that \textcolor{black}{the} existing self-supervised methods are susceptible to the quality of synthetic data. To improve the robustness of \textcolor{black}{the} prior arts, we have proposed a simple yet effective framework\textcolor{black}{, named} Discrepancy Aware Framework (DAF). DAF introduces the teacher-student \textcolor{black}{model} to yield the discrepancy map and serves it as an additional cue that reveals the location of anomalies to the segmentation decoder, alleviating the decoder’s reliance on the appearance of synthetic data. Meanwhile, the complement of the discrepancy map for segmentation contributes significantly to robustness as well. Extensive experiments have shown that DAF surpasses previous self-supervised methods significantly when faced with simple synthetic strategies in anomaly detection and localization, demonstrating its excellent robustness.


{
\small
\bibliographystyle{IEEEtran}
\bibliography{egbib}
}

\vfill

\end{document}